# The Algonauts Project 2023 Challenge: How the Human Brain Makes Sense of Natural Scenes

Alessandro T. Gifford[1,*], Benjamin Lahner[2], Sari Saba-Sadiya[3], Martina G. Vilas[3], Alex Lascelles[2], Aude Oliva[2], Kendrick Kay[4], Gemma Roig[3,5], Radoslaw M. Cichy[1,*]

[1] Department of Education and Psychology, Freie Universität Berlin, Germany
[2] Computer Science and Artificial Intelligence Laboratory, MIT, USA
[3] Department of Computer Science, Goethe Universität Frankfurt, Germany
[4] Center for Magnetic Resonance Research (CMRR), University of Minnesota, USA
[5] Hessian Center for AI (hessian.AI), Darmstadt, Germany

The sciences of biological and artificial intelligence are ever more intertwined. Neural computational principles inspire new intelligent machines, which are in turn used to advance theoretical understanding of the brain. To promote further exchange of ideas and collaboration between biological and artificial intelligence researchers, we introduce the 2023 installment of the Algonauts Project challenge: How the Human Brain Makes Sense of Natural Scenes (http://algonauts.csail.mit.edu). This installment prompts the fields of artificial and biological intelligence to come together towards building computational models of the visual brain using the largest and richest dataset of fMRI responses to visual scenes, the Natural Scenes Dataset (NSD). NSD provides high-quality fMRI responses to ~73,000 different naturalistic colored scenes, making it the ideal candidate for data-driven model building approaches promoted by the 2023 challenge. The challenge is open to all and makes results directly comparable and transparent through a public leaderboard automatically updated after each submission, thus allowing for rapid model development. We believe that the 2023 installment will spark symbiotic collaborations between biological and artificial intelligence scientists, leading to a deeper understanding of the brain through cutting-edge computational models and to novel ways of engineering artificial intelligent agents through inductive biases from biological systems.

## Introduction

In the last decade the deep learning revolution has profoundly impacted scientific research endeavors (Sejnowski, 2018; Baldi, 2021). During this time the quest for solving intelligence in both its artificial and biological form has made remarkable progress: deep learning algorithms, originally inspired by the visual system of the mammalian brain (Fukushima & Miyake, 1982), are now both state-of-the-art AI agents (Krizhevsky et al., 2017) and scientific models of the brain itself (Yamins & DiCarlo, 2016; Cichy & Kaiser, 2019; Kietzmann et al., 2019; Richards et al., 2019; Saxe et al., 2021). Thus, research in artificial and biological intelligence is ever more intertwined. Furthermore, this synergy is accelerated by the release of large neural datasets that allow training deep learning models end-to-end on brain data (Allen et al., 2022; Gifford et al., 2022), and by challenges that exploit these and other neural datasets to develop better models of intelligence (Schrimpf et al., 2020; Willeke et al., 2022).

We contribute to the symbiosis between biological and artificial intelligence with the third installment of the Algonauts Project challenge, titled "How the Human Brain Makes Sense of Natural Scenes". The 2023 installment continues the spirit of the 2019 and 2021 editions of the Algonauts Project in its goal of predicting human visual brain responses through computational models (Cichy et al., 2019; Cichy et al., 2021). Yet, it goes beyond the previous challenges in that it is based on the largest and richest dataset of neural responses to natural scenes, the Natural Scenes Dataset (NSD) (Allen et al., 2022). We focus on visual scene understanding since vision is an unresolved problem in the sciences of artificial and biological intelligence alike (Szegedy et al, 2014; Gheiros et al., 2019; DiCarlo et al., 2012), and one of the areas where collaboration between these two fields has been most fruitful (Yamins & DiCarlo, 2016; Li et al., 2019; Safarani et al., 2021; Dapello et al., 2022).

We believe that the Algonauts Project will lead to significant advances in both the understanding of the brain through artificial intelligence models (Yamins & DiCarlo, 2016; Cichy and Kaiser, 2019; Kietzmann et al., 2019; Richards et al., 2019) and the engineering of better AI agents through biological intelligence constraints (Hassabis et al., 2017; Sinz et al., 2019; Ullman, 2019; Yang et al., 2022; Toneva & Wehbe, 2019; Li et al., 2019; Safarani et al., 2021; Dapello et al., 2022), thus contributing to the ever stronger symbiosis between the sciences of biological and artificial intelligence.

* **Correspondence:** alessandro.gifford@gmail.com; rmcichy@zedat.fu-berlin.de

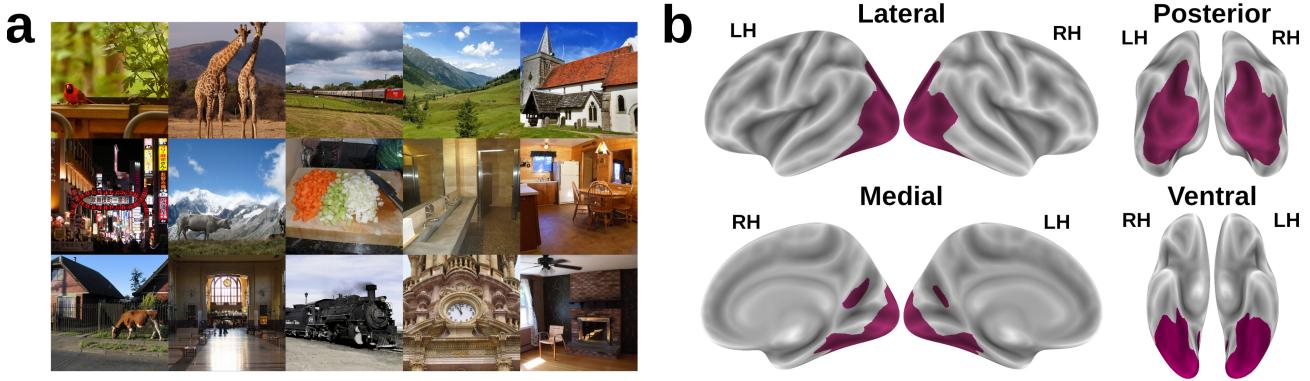

**Figure 1.** Challenge data. (**a**) Exemplar NSD stimuli images. All images consist of natural scenes taken from the COCO database. (**b**) Depiction of the cortical surface vertices used in the challenge (purple).

## Materials and Methods

**Challenge goal.** The goal of the Algonauts 2023 Project challenge is to promote the development of cutting-edge encoding models that predict neural responses to visual stimuli, and to provide a common platform that catalyzes collaborations between the fields of biological and artificial intelligence.

**Primer on encoding models.** Encoding models are algorithms that predict how the brain responds to (i.e., encodes) certain stimuli (Naselaris et al., 2011; van Gerven, 2017). In visual neuroscience, an encoding model typically consists in an algorithm that takes image pixels as input, transforms them into model features, and maps these features onto brain data (e.g., fMRI activity), effectively predicting the neural responses to images.

**Challenge data.** In this challenge participants will leverage the unprecedented size of NSD (Allen et al., 2022) to build encoding models of the visual brain. NSD is a massive 8-subject dataset of high-quality 7T fMRI responses to ~73,000 different natural scenes (presented during central fixation) from the Common Objects in Context (COCO) database (Lin et al., 2014) (**Fig. 1a**). The challenge uses preprocessed fMRI responses (BOLD response amplitudes) from each subject that have been projected onto a common cortical surface group template (FreeSurfer's fsaverage surface). Brain surfaces are composed of vertices, and the challenge data consists of a subset of cortical surface vertices in the visual cortex (a region of the brain specialized in processing visual input) that were maximally responsive to visual stimulation (**Fig. 1b**). We provide the data in right and left hemispheres. Further information on NSD acquisition and preprocessing is provided on the challenge website[1] and in the NSD paper[2]. Since model building requires independent data splits for training and testing, we partitioned the challenge data into non-overlapping train and test splits coming from, respectively, the publicly released NSD data and the last three NSD scan sessions from each subject (which are not publicly available).

**Train split.** For each of the 8 subjects we provide [9841, 9841, 9082, 8779, 9841, 9082, 9841, 8779] images, along with the corresponding fMRI visual responses, z-scored within each NSD scan session and averaged across image repeats. These data can be used to train encoding models.

**Test split.** For each of the 8 subjects we provide [159, 159, 293, 395, 159, 293, 159, 395] images (different from the train images) but withhold the corresponding fMRI visual responses. Challenge participants are asked to use their encoding models to predict the fMRI responses of these test images.

**ROI indices.** The visual cortex is divided into multiple areas having different functional properties, referred to here as regions-of-interest (ROIs). Along with the fMRI data, we provide ROI indices for selecting vertices belonging to specific visual ROIs; challenge participants can optionally use these ROI indices at their own discretion (e.g., to build different encoding models for functionally different regions of the visual cortex). However, the challenge evaluation metric is computed over all available vertices, not over any single ROI.

**Development kit.** We provide a Colab tutorial[3] in Python where we show how to: (**i**) load and visualize the fMRI data, its ROIs and the corresponding stimulus images; (**ii**) build and evaluate linearizing encoding models (Naselaris et al., 2011) using a pretrained AlexNet (Krizhevsky, 2014) architecture; (**iii**) prepare predicted brain responses of the test images in the right format for challenge submission.

---

[1] http://algonauts.csail.mit.edu
[2] https://www.nature.com/articles/s41593-021-00962-x
[3] https://colab.research.google.com/drive/1bLJGP3bAo_hAOwZPHpiSHKlt97X9xsUw?usp=sharing

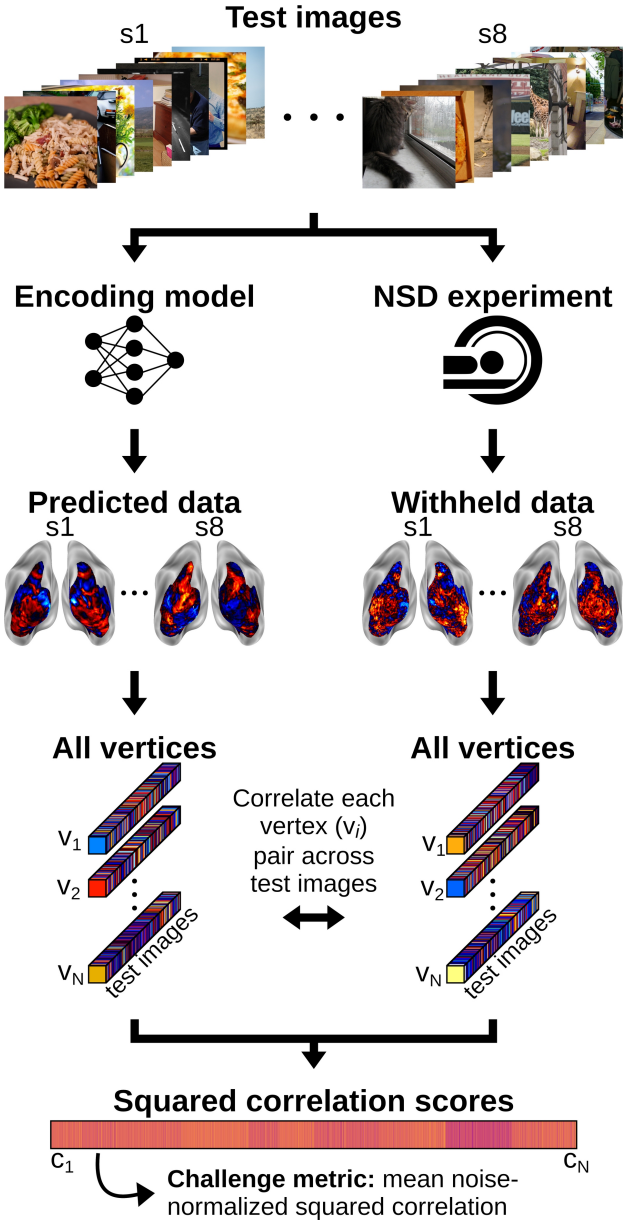

**Figure 2.** Challenge evaluation metric. Once participants submit their predictions of the fMRI responses to the test images of all 8 subjects (s1, …, s8), we evaluate prediction accuracy using the withheld fMRI test data. In detail, we concatenate the predicted data vertices ($V_1$, $V_2$, …, $V_N$) of all subjects and hemispheres, correlate them with the corresponding withheld data vertices (across the test stimuli images), and square the correlations, resulting in one squared correlation score ($C_1$, …, $C_N$) for each vertex. The challenge evaluation metric is the mean noise-normalized squared correlation score across all vertices

**Challenge submission and evaluation metric.** To quantify accuracy of encoding models, participants submit their fMRI predictions for the test images. For each NSD subject and hemisphere, we will correlate the fMRI predicted data with the corresponding ground truth (withheld) data at all vertices (across images), square the resulting correlation scores and normalize them with respect to the noise ceiling (reflecting the total predictable variance given the level of noise in the data). The resulting values will indicate how much of the predictable variance has been accounted for by the models. The overall challenge evaluation metric, which quantifies the performance of each participant's submission, is the mean noise-normalized squared correlation score over all vertices from all subjects (**Fig. 2**):

$$\text{metric} = Mean\left\{\frac{R_1^2}{NC_1}, \dots, \frac{R_v^2}{NC_v}\right\} \times 100$$

$$R_v = \text{corr}(G_v, P_v) = \\ = \frac{\sum_t (G_{v,t} - \bar{G}_v)(P_{v,t} - \bar{P}_v)}{\sqrt{\sum_t (G_{v,t} - \bar{G}_v)^2 \sum_t (P_{v,t} - \bar{P}_v)^2}},$$

where $v$ is the index of vertices (over all subjects and hemispheres), $t$ is the index of the test stimuli images, $G$ and $P$ correspond to, respectively, the ground truth and predicted fMRI test data, $\bar{G}$ and $\bar{P}$ are the ground truth and predicted fMRI test data averaged across test stimuli images, $R$ is the Pearson correlation coefficient between $G$ and $P$, and $NC$ is the noise ceiling.

**Baseline model.** The baseline model score of the challenge reflects a linearizing encoding model (Naselaris et al., 2011) built using a pretrained AlexNet. Its mean noise-normalized prediction accuracy over all subjects, hemispheres and vertices is 40.48% of the total predictable variance.

**Rules.** To encourage broad participation, the challenge has a simple submission process with minimal rules. Participants can use any encoding model derived from any source and trained on any type of data, and can make a limited number of submissions per day (the leaderboard is automatically updated after each submission). However, using the test split for training (in particular brain data generated using the test images) is prohibited. To promote open science, participants who wish to be considered for the challenge contest will have to submit a report to a preprint server describing their encoding algorithm. Furthermore, the top three entries are required to make their code openly available and, along with other prizes, will have the chance to present their winning encoding models through a talk at the Cognitive Computational Neuroscience (CCN) conference in 2023.

## Discussion

**Relation to similar initiatives.** The Algonauts project relates to initiatives such as Brain-Score (Schrimpf et al., 2020) and Sensorium 2022 (Willeke et al., 2022), which also establish benchmarks and leaderboards. However, the Algonauts Project differs from these complementary efforts by emphasizing human data, by focusing on colored images of complex naturalistic scenes including multiple object concepts, by leveraging a whole-brain fMRI dataset that extensively samples stimulus variation (Allen et al., 2022; Naselaris et al., 2021), and by incorporating educational components (hands-on computational modeling tutorial; talks by the top three challenge winners; panel discussion on challenges) at a dedicated session at CCN in 2023.

**Prediction vs. explanation.** Having a model that perfectly predicts a phenomenon does not by itself explain the phenomenon. However, prediction and explanation are related goals and complement each other (Cichy & Kaiser, 2019). First, successful explanations must also provide successful predictions (Breiman, 2001; Yarkoni & Westfall, 2017). Second, prediction accuracy can shed light on properties that make particular models successful, allowing testing or generating hypotheses and guiding future engineering steps. Finally, predictive success as an evaluation criterion circumvents the challenges of evaluation on qualitative or subjective grounds.

**From artificial to biological intelligence.** During the last decade, interactions between biological and artificial intelligence have profoundly affected neuroscientific discovery, and machine/deep learning algorithms have become state-of-the-art models of the brain (Yamins & DiCarlo, 2016; Cichy & Kaiser, 2019; Kietzmann et al., 2019; Richards et al., 2019; Saxe et al., 2021). However, due to their large parameter number, these algorithms require massive amounts of data to properly train (Russakovsky et al., 2015). We addressed this by basing the 2023 installment of the Algonauts Project on the NSD dataset (Allen et al., 2022). The unprecedented scale of NSD, along with its extensive sampling of stimulus variation, allows for data-driven model building approaches such as training deep learning architectures end-to-end to predict neural responses to visual stimuli (Allen et al., 2022; St-Yves et al., 2022; Khosla & Wehbe, 2022; Gifford et al., 2022). Directly infusing deep learning models with brain data enables a novel type of interaction between biological and artificial intelligence, which in our opinion will catalyze breakthroughs in neuroscientific research.

**From biological to artificial intelligence.** Artificial intelligence too can benefit from interactions with biological intelligence. Biological systems constitute a proof of principle for how a complex computational problem can be solved, and thus can guide the engineering of new artificial intelligence models (Hassabis et al., 2017; Sinz et al., 2019; Ullman, 2019; Yang et al., 2022). This research direction is especially promising for improving artificial agents in domains at which biological agents excel (e.g., out-of-domain generalization, transfer learning, adversarial robustness, few-shot learning), and even for endowing artificial agents with cognitive faculties idiosyncratic to humans (e.g., planning, creativity, imagination). Artificial intelligence researchers have been successfully exploring these possibilities for decades. As an example, the structure of the current state-of-the-art artificial intelligence algorithms, deep neural networks, has been inspired by the structure of the visual system of the mammalian brain (Fukushima & Miyake, 1982). Furthermore, a growing amount of literature has started to exploit neural data representations to train natural language processing and computer vision algorithms, resulting in models with improved performance and adversarial robustness (Toneva & Wehbe, 2019; Li et al., 2019; Safarani et al., 2021; Dapello et al., 2022). The Algonauts Project fosters this exciting area of research by promoting interactions between the fields of artificial and biological intelligence.

**The future of the project.** We hope that the 2023 installment of the Algonauts Project will continue to inspire new challenges and collaborations at the intersection of artificial and biological intelligence sciences. We believe that both communities will benefit from jointly tackling open questions on how perception and cognition are solved in brains and machines. We welcome researchers interested in initiating similar initiatives or collaborating with the Algonauts Project to contribute ideas and datasets.

## Acknowledgments

This research was funded by DFG (CI-241/1-1, CI241/1-3,CI-241/1-7) and ERC grant (ERC-2018-StG) to RMC; NSF award (1532591) in Neural and Cognitive Systems, the Vannevar Bush Faculty Fellowship program funded by the ONR (N00014-16-1-3116) and MIT-IBM Watson AI Lab to AO; the Alfons and Gertrud Kassel foundation to GR. Collection of the NSD dataset was supported by NSF IIS-1822683 and NSF IIS-1822929.

## References

Allen EJ, St-Yves G, Wu Y, Breedlove JL, Prince JS, Dowdle LT, Nau M, Caron B, Pestilli F, Charest I, Hutchinson JB, Naselaris T, Kay K. 2022. A massive 7T fMRI dataset to bridge cognitive neuroscience and computational intelligence. *Nature Neuroscience*,

25(1):116–126.

Baldi P. 2021. Deep learning in science. *Cambridge University Press*.

Breiman L. 2001. Statistical modeling: The two cultures. *Statistical science*, 16(3):199-231.

Cichy RM, Dwivedi K, Lahner B, Lascelles A, Iamshchinina P, Graumann M, Andonian A, Murty NAR, Kay K, Roig G, Oliva A. 2021. The Algonauts Project 2021 Challenge: How the Human Brain Makes Sense of a World in Motion. *arXiv preprint*, arXiv:2104.13714.

Cichy RM, Kaiser D. 2019. Deep neural networks as scientific models. *Trends in cognitive sciences*, 23(4):305-317.

Cichy RM, Roig G, Andonian A, Dwivedi K, Lahner B, Lascelles A, Mohsenzadeh Y, Ramakrishnan K, Oliva A. 2019. The algonauts project: A platform for communication between the sciences of biological and artificial intelligence. *arXiv preprint*, arXiv:1905.05675.

Dapello J, Kar K, Schrimpf M, Geary R, Ferguson M, Cox DD, DiCarlo JJ. 2022. Aligning Model and Macaque Inferior Temporal Cortex Representations Improves Model-to-Human Behavioral Alignment and Adversarial Robustness. *bioRxiv*.

DiCarlo JJ, Zoccolan D, Rust NC. 2012. How does the brain solve visual object recognition? *Neuron*, 73(3):415-434.

Fukushima K, Miyake S. 1982. Neocognitron: A self-organizing neural network model for a mechanism of visual pattern recognition. *Competition and cooperation in neural nets*, 267-285.

Geirhos R, Rubisch P, Michaelis C, Bethge M, Wichmann FA, Brendel W. 2018. ImageNet-trained CNNs are biased towards texture; increasing shape bias improves accuracy and robustness. *arXiv preprint*, arXiv:1811.12231.

Gifford AT, Dwivedi K, Roig G, Cichy RM. 2022. A large and rich EEG dataset for modeling human visual object recognition. *NeuroImage*, 119754.

Hassabis D, Kumaran D, Summerfield C, Botvinick M. 2017. Neuroscience-inspired artificial intelligence. *Neuron*, 95(2):245-258.

Khosla M, Wehbe L. 2022. High-level visual areas act like domain-general filters with strong selectivity and functional specialization. *bioRxiv*.

Kietzmann TC, McClure P, Kriegeskorte N. 2019. Deep Neural Networks in Computational Neuroscience. *Oxford Research Encyclopedia of Neuroscience*.

Krizhevsky A. 2014. One weird trick for parallelizing convolutional neural networks. *arXiv preprint*, arXiv:1404.5997.

Krizhevsky A, Sutskever I, Hinton GE. 2017. Imagenet classification with deep convolutional neural networks. *Communications of the ACM*, 60(6):84-90.

Li Z, Brendel W, Walker E, Cobos E, Muhammad T, Reimer J, Bethge M, Sinz F, Pitkow Z, Tolias A. 2019. Learning from brains how to regularize machines. *Advances in neural information processing systems*, 32.

Lin TY, Maire M, Belongie S, Hays J, Perona P, Ramanan D, Dollár P, Zitnick CL. 2014. Microsoft coco: Common objects in context. *European conference on computer vision*, 740-755.

Naselaris T, Allen E, Kay K. 2021. Extensive sampling for complete models of individual brains. *Current Opinion in Behavioral Sciences*, 40:45-51.

Naselaris T, Kay KN, Nishimoto S, Gallant JL. 2011. Encoding and decoding in fMRI. *Neuroimage*, 56(2):400-410.

Richards BA, Lillicrap TP, Beaudoin P, Bengio Y, Bogacz R, Christensen A, … , Kording KP. 2019. A deep learning framework for neuroscience. *Nature neuroscience*, 22(11):1761-1770.

Russakovsky O, Deng J, Su H, Krause J, Satheesh S, Ma S, Huang Z, Karpathy A, Khosla A, Bernstein M, Berg AC, Fei-Fei L. 2015. ImageNet large scale visual recognition challenge. *International journal of computer vision*, 115(3):211-252.

Safarani S, Nix A, Willeke K, Cadena S, Restivo K, Denfield G, Tolias A, Sinz F. 2021. Towards robust vision by multi-task learning on monkey visual cortex. *Advances in Neural Information Processing Systems*, 34:739-751.

Saxe A, Nelli S, Summerfield C. 2021. If deep learning is the answer, what is the question? *Nature Reviews Neuroscience*, 22(1):55-67.

Schrimpf M, Kubilius J, Lee MJ, Murty NAR, Ajemian R, DiCarlo JJ. 2020. Integrative benchmarking to advance neurally mechanistic models of human intelligence. *Neuron*, 108(3):413-423.

Szegedy C, Zaremba W, Sutskever I, Bruna J, Erhan D, Goodfellow I, Fergus R. 2013. Intriguing properties of neural networks. *arXiv preprint*, arXiv:1312.6199.

Sejnowski TJ. 2018. The deep learning revolution. *MIT press*.

Sinz FH, Pitkow X, Reimer J, Bethge M, Tolias AS. 2019. Engineering a less artificial intelligence. *Neuron*, 103(6):967-979.

St-Yves G, Allen EJ, Wu Y, Kay K, Naselaris T. 2022. Brain-optimized neural networks learn non-hierarchical models of representation in human visual cortex. *bioRxiv*.

Toneva M, Wehbe L. 2019. Interpreting and improving natural-language processing (in machines) with natural language-processing (in the brain). *Advances in Neural Information Processing Systems*, 32.

Ullman S. 2019. Using neuroscience to develop artificial intelligence. *Science*, 363(6428):692-693.

van Gerven MA. 2017. A primer on encoding models in sensory neuroscience. *Journal of Mathematical Psychology*, 76:172-183.

Willeke KF, Fahey PG, Bashiri M, Pede L, Burg MF, Blessing C, … , Sinz FH. 2022. The Sensorium competition on predicting large-scale mouse primary visual cortex activity. *arXiv preprint, arXiv:2206.08666.*

Yamins DL, DiCarlo JJ. 2016. Using goal-driven deep learning models to understand sensory cortex. *Nature neuroscience*, 19(3):356-365.

Yang X, Yan J, Wang W, Li S, Hu B, Lin J. 2022. Brain-inspired models for visual object recognition: an overview. *Artificial Intelligence Review*, 1-49.

Yarkoni T, Westfall J. 2017. Choosing prediction over explanation in psychology: Lessons from machine learning. *Perspectives on Psychological Science*, 12(6):1100-1122.